\pdfoutput=1
\documentclass[11pt]{article}

\usepackage[final]{acl}

\usepackage{times}
\usepackage{latexsym}

\usepackage[T1]{fontenc}
\usepackage[utf8]{inputenc}

\usepackage{microtype}

\usepackage{inconsolata}

\usepackage{mathtools, nccmath}
\usepackage{graphicx}
\graphicspath{ {imgs} }
\usepackage{booktabs}
\usepackage{url}

\title{Reinforcement Learning for Edit-Based Non-Autoregressive\\Neural Machine Translation}

\author{Hao Wang$^{1}$\thanks{Work done during internship at CyberAgent AI Lab.}\qquad Tetsuro Morimura$^2$\qquad Ukyo Honda$^2$\qquad Daisuke Kawahara$^1$\\
$^1$ Waseda University\qquad
$^2$ CyberAgent\\
$^1$ {\texttt{\{conan1024hao@akane., dkw@\}waseda.jp}}\\
$^2$ {\texttt{\{morimura\_tetsuro, honda\_ukyo\}@cyberagent.co.jp}}\\
}

\begin{document}
\maketitle
\begin{abstract}
Non-autoregressive (NAR) language models are known for their low latency in neural machine translation (NMT).
However, a performance gap exists between NAR and autoregressive models due to the large decoding space and difficulty in capturing dependency between target words accurately.
Compounding this, preparing appropriate training data for NAR models is a non-trivial task, often exacerbating exposure bias. 
To address these challenges, we apply reinforcement learning (RL) to Levenshtein Transformer, a representative edit-based NAR model, demonstrating that RL with self-generated data can enhance the performance of edit-based NAR models.
We explore two RL approaches: stepwise reward maximization and episodic reward maximization.
We discuss the respective pros and cons of these two approaches and empirically verify them.
Moreover, we experimentally investigate the impact of temperature setting on performance, confirming the importance of proper temperature setting for NAR models' training.
\end{abstract}

\section{Introduction}
Non-autoregressive (NAR) language models~\citep{gu2018nonautoregressive} generate translations in parallel, enabling faster inference and having the potential for real-time translation applications.
However, despite their computational efficiency, NAR models have been observed to underperform autoregressive (AR) models due to the challenges posed by the large decoding space and difficulty in capturing dependency between target words accurately~\citep{gu2018nonautoregressive}.
To bridge the performance gap, many NAR architectures and training methods have been proposed, including edit-based models like Insertion Transformer~\citep{stern2019insertion} and Levenshtein Transformer~\citep{gu2019levenshtein}.
Prior research has also explored knowledge distillation~\citep{ghazvininejad-etal-2019-mask}, which is effective but introduces additional complexity.

Unlike AR models, preparing teacher data and designing appropriate training objectives have always been challenging for NAR models~\citep{li2023diffusion}.
Teacher forcing with inappropriate teacher data may exacerbate the exposure bias problem~\citep{ranzato2016sequence}, affecting model performance.
Reinforcement learning (RL) is known for its ability to tackle the exposure bias~\citep{ranzato2016sequence} and alleviate the object mismatch issue~\citep{ding2017coldstart}.
Despite its importance, explorations of RL for NAR are still scarce.
\citet{shao-etal-2021-sequence} proposed a method for reducing the estimation variance.
However, this method is only applicable to NAR models with a fixed output length, which is unsuitable for edit-based models.

In this paper, we empirically analyze conditions for performance improvement in applying RL to edit-based NAR models in neural machine translation (NMT).
Specifically, we focus on Levenshtein Transformer (LevT)~\citep{gu2019levenshtein}, a prominent edit-based NAR architecture that has shown promise in reducing decoding latency and flexible length adjustment.
We demonstrate that RL with self-generated data significantly improves LevT's performance.
Importantly, our methods are orthogonal to existing research on NAR architectures, indicating potential for widespread applicability.
We explore two RL approaches: stepwise reward maximization, which computes rewards after each edit operation, and episodic reward maximization, which only computes rewards after all generations are completed. 
We analyze these two approaches' respective advantages and disadvantages and empirically verify them.
Furthermore, through a series of experiments, we investigate the impact of temperature settings on softmax sampling, aiming to identify the optimal temperature that strikes a balance between exploration and exploitation during the RL training process.
%=================================================================================================

\section{Background}

\paragraph{Reinforcement Learning}
Reinforcement learning has been widely applied to improve the performance of AR NMT models~\citep{ranzato2016sequence, bahdanau2016actor, wu2016googles} because its ability to train models to optimize non-differentiable score functions and tackle the exposure bias problem~\citep{ranzato2016sequence}.
In practice, REINFORCE~\citep{williams1992simple} with a baseline is commonly used for estimating the policy gradient, which can be computed as follows:
\begin{equation}
\label{rl}
    \bigtriangledown_{\theta} L(\theta) \approx -(r(y) - b(s)) \bigtriangledown_{\theta} \mathrm{log} \pi_{\theta}(y|s),
\end{equation}
where $r$ is the reward function, $b$ is the baseline, $y$ is a sample from policy $\pi_{\theta}$ and state $s$.
%-------------------------------------------------------------------------------------------------

% \paragraph{Softmax with Temperature}
% In the domain of RL, we need to consider the exploration-exploitation trade-off~\citep{Sutton2018rl}, where temperature $\tau$ is an important parameter.
% $\tau$ is used to control the softness of the softmax distribution.
% Bigger $\tau$ results in larger randomness, while smaller $\tau$ results in more decisive samplings.
% After applying the temperature $\tau$, softmax's formula becomes
% \begin{equation}
%     p_i = \frac{\mathrm{exp}(y_i / \tau)}{\sum_i \mathrm{exp}(y_i / \tau)}.
% \end{equation}
% \citet{kiegeland-kreutzer-2021-revisiting} shows that training with an increased temperature can mitigate the peakiness effect due to RL~\citep{Choshen2020On}, indicating that a suitable temperature is significant for RL training in NMT.

\paragraph{Softmax with Temperature}
In the domain of RL, we need to consider the exploration-exploitation trade-off~\citep{Sutton2018rl}, where temperature $\tau$ is an important parameter.
$\tau$ is used to control the softness of the softmax distribution,
\begin{equation}
    p_i = \frac{\mathrm{exp}(y_i / \tau)}{\sum_i \mathrm{exp}(y_i / \tau)}.
\end{equation}
A larger $\tau$ leads to a more uniform distribution, promoting exploration, while a smaller $\tau$ creates a more peaky distribution, emphasizing exploitation.

\citet{kiegeland-kreutzer-2021-revisiting} shows that training with an increased temperature can mitigate the peakiness effect due to RL~\citep{Choshen2020On}, indicating that a suitable temperature is significant for RL training in NMT.
%-------------------------------------------------------------------------------------------------

\paragraph{RL for NAR}
Compared to AR methods, studies of reinforcement learning for NAR remain unexplored.
\citet{shao-etal-2021-sequence} proposed a method to reduce the estimation variance of REINFORCE by fixing the predicted word at position $t$ and sampling words of other positions for $n$ times.
However, this method is only applicable to models with a fixed length, which is unsuitable for edit-based models.
%-------------------------------------------------------------------------------------------------

\begin{figure}[t]
    \centering
    \includegraphics[width=1\linewidth]{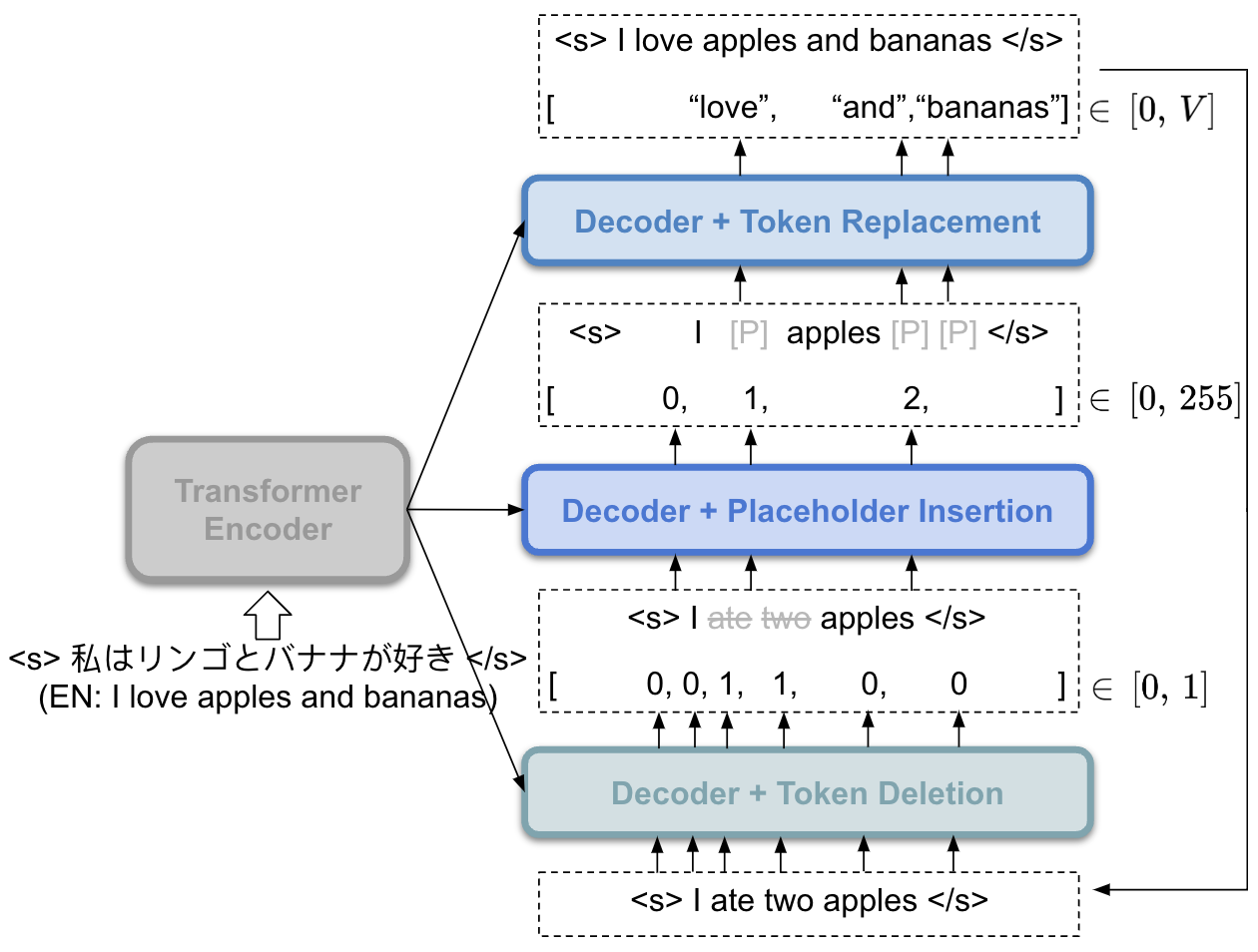}
    \caption{The illustration of Levenshtein Transformer's decoding process~\citep{gu2019levenshtein}. In each decoding iteration, three edit operations are performed sequentially: delete tokens, insert placeholders, and replace placeholders with new tokens.}
    \label{fig:levt}
    \vspace{-10pt}
\end{figure}

\paragraph{Levenshtein Transformer}
Levenshtein Transformer~\citep{gu2019levenshtein} is an NAR model based on three edit operations: delete tokens, insert placeholders, and replace placeholders with new tokens.
It uses a supervised dual-policy learning algorithm to minimize the Levenshtein distance~\citep{Levenshtein1965BinaryCC} for training and greedy sampling for decoding.
The decoding stops when two consecutive refinement iterations return the same output or a maximum number of iterations (set to 10) is reached.
We illustrate the decoding process in Figure~\ref{fig:levt}.

% LevTは適当なルールでcorruptして教師データを作成しているけど、それはdeconding時の文章との一致性があるか、バイアスがあるかは不明であって、教師データの作成が難しい、とか
LevT's dual-policy learning generates teacher data by corrupting the ground truth and reconstructing it with its adversary policy.
This mechanism not only offers a unique approach to data generation but also underscores the inherent difficulty in preparing teacher data.
This introduces concerns regarding the exposure bias, particularly whether the training process can maintain consistency with the text during decoding.
To address this issue, we employ RL approaches that use self-generated data for training.
%=================================================================================================
\section{Approaches}
In this section, we present our reinforcement learning approaches in detail.
We train a Levenshtein Transformer model as our baseline using the dual-policy learning algorithm.
Based on it, we introduce two distinct RL approaches within the REINFORCE framework: stepwise reward maximization and episodic reward maximization.
Moreover, we present our methods for temperature control.
%-------------------------------------------------------------------------------------------------
\begin{figure*}[t]
    \centering
    \includegraphics[width=1\linewidth]{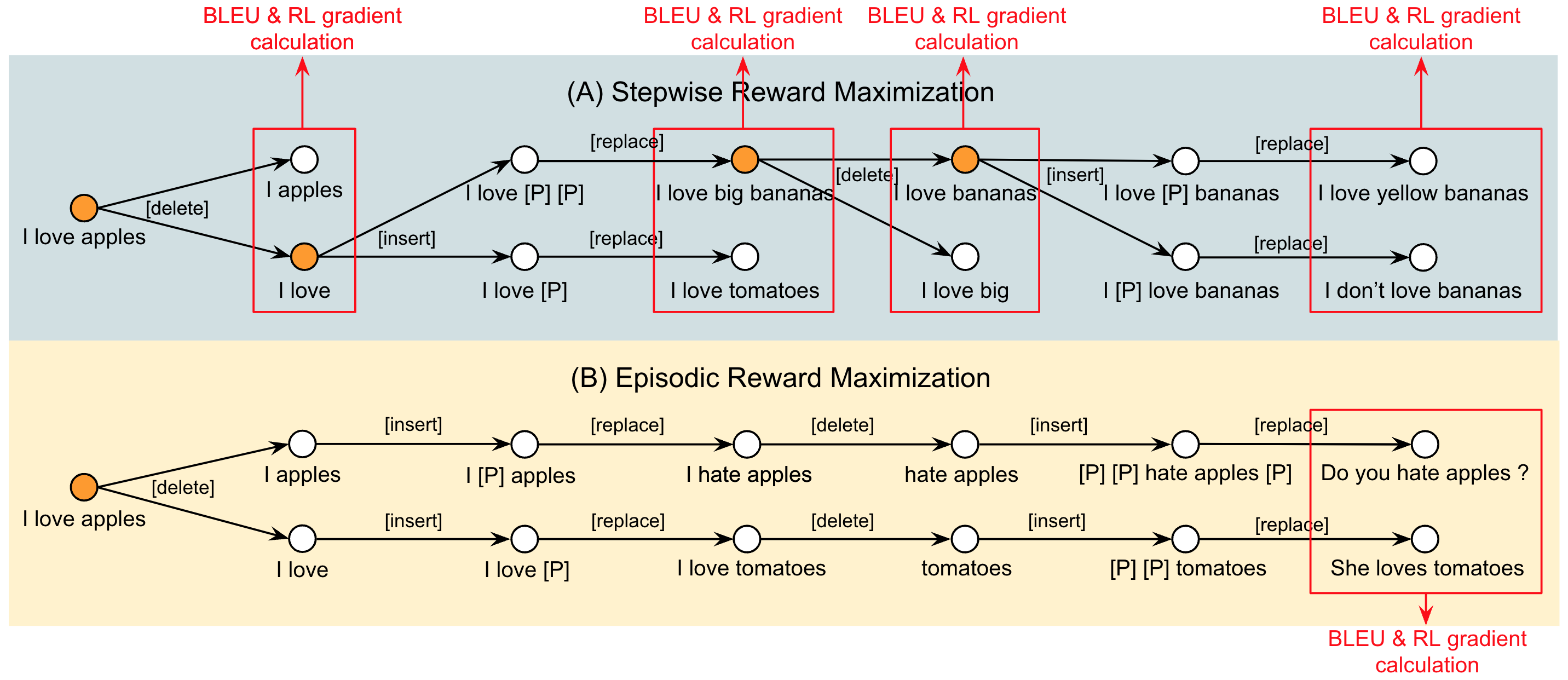}
    \vspace{-25pt}
    \caption{The illustration of the two RL approaches. (A) is the stepwise reward maximization, which randomly samples from a previous node for each edit operation and calculates BLEU and RL gradient after each edit operation (except for the insert operation, since it is not easy to calculate BLEU after inserting placeholders). (B) is the episodic reward maximization, where each sample is edited multiple times in a linear fashion, without branching into different paths, and BLEU and RL gradient are calculated only after the completion of all edit operations. At every orange node, we sample $k$ times from this node (in this example, the sample size $k$ is 2).}
    \label{fig:method}
    \vspace{-10pt}
\end{figure*}
%-------------------------------------------------------------------------------------------------
\paragraph{Stepwise Reward Maximization}
General RL training methods for AR NMT models are all episodic\footnote{In this context, ``episodic'' denotes training based on entirely generated sequences}, as it is difficult to calculate BLEU~\citep{papineni-etal-2002-bleu} when the sentence is not fully generated.
In contrast, NAR models can calculate BLEU on outputs at each decoding step.
From the perspective of estimating a more accurate gradient, we propose stepwise reward maximization, which calculates reward for each edit operation\footnote{In practice, since it is not easy to calculate BLEU after inserting placeholders, we consider placeholder insertion and token replacement as one edit operation.} using score differences from one previous edit.
Since every step's reward is calculated separately, this approach should be easier to learn than episodic approaches~\citep{Sutton2018rl}.
However, it is also more prone to learning bias since the editing process is inherently multi-step.
This drawback should not be emphasized since maximizing the reward for each step will likely maximize the episodic reward in NAR models' training.

We use a leave-one-out baseline~\citep{luo2020better} for $b(s)$ in Equation~\ref{rl} instead of the greedy baseline proposed in SCST~\citep{Rennie_2017_CVPR} because the greedy decoding is too strong in LevT, which makes gaining positive rewards in SCST difficult and may reduce learning efficiency.
For each edit, we sample $k$ actions from the policy at this point.
Then, we calculate the baseline as follows:
\begin{equation}
    b_i(s) = \frac{1}{k - 1}\sum_{j \neq i}r(y_j),
\end{equation}
where $y_j$ is the $j$th sample from the current policy.
The final RL gradient estimation becomes
\begin{equation}\label{leave-one-out}
    \bigtriangledown_{\theta} L(\theta) \approx -(r(y_i) - b_i(s)) \bigtriangledown_{\theta} \mathrm{log} \pi_{\theta}(y_i|s).
\end{equation}

In a straightforward implementation, one might consider applying sampling again to all $k$ samples from the last edit.
However, this will cause a combination explosion when the number of edit operations increases.
Practically, we randomly choose a sample from the previous edit to perform the subsequent operations.
We show an illustration of the sampling process in (A) of Figure~\ref{fig:method} and pseudo code of our algorithm in Appendix~\ref{pseudo-stepwise}.
%-------------------------------------------------------------------------------------------------
\paragraph{Episodic Reward Maximization}
We also introduce episodic reward maximization, which calculates rewards only once for each sample and gives all actions the same weight.
It is a more traditional way to train NMT models in RL.
It allows unbiased learning but may not be efficient.

We use the leave-one-out baseline for the episodic reward as well as the stepwise reward.
We sample $k$ samples from the initial input.
Each sample will be edited multiple times without a branch.
After the final edit, we calculate the rewards and baselines.
We show an illustration of the sampling process in (B) of Figure~\ref{fig:method} and pseudo code of our algorithm in Appendix~\ref{pseudo-episodic}.
%-------------------------------------------------------------------------------------------------
\paragraph{Temperature Control}
Applying RL to NAR differs significantly from AR because there could be various types of actions rather than just predicting the next token, like deletion and insertion.
% Further exploration is required for temperature control during NAR models' training. 
Due to this difficulty, NAR may need more fine-grained temperature control during training.
To investigate the impact of exploration and exploitation in the training process, we explore five different settings of the temperature.
Due to the large decoding space of Levenshtein Transformer, default temperature 1 may result in poor rewards, and too small temperature may result in peaky distribution, which are both harmful to learning.
We use three constant temperature settings set to 0.1, 0.5, and 1 to verify the effect of temperature magnitude.

An annealing schedule is known for balancing the trade-off between model accuracy and variance during training~\citep{jang2016categorical}.
There are two ways of thinking here.
First, to reduce the exposure bias, we want to get close to the decoding scenario, which is greedy decoding in our experiments.
Thus, we can apply a regular annealing schedule to gradually reduce the temperature from 1 to 0.1 during training.
The temperature function can be written as follows:
\begin{equation}
\tau_{i+1} = \max(\tau_i * \mathrm{exp}(-\frac{\mathrm{log}(\tau_0 / \tau_T)}{T}), \tau_T),
\end{equation}
where $T$ is the number of total training steps, and $\tau_0$ and $\tau_T$ are the initial and the target temperatures.

Second, using high temperatures in the early stages of training may lead to poor rewards and result in low learning efficiency.
We can apply an inverted annealing schedule to gradually increase the temperature from 0.1 to 1, guaranteeing stable training in the early stages and gradually increasing the exploration space for efficient training.
The temperature function can be written as follows:
\begin{equation}
\tau_{i+1} = \min(\tau_i / \mathrm{exp}(-\frac{\mathrm{log}(\tau_T / \tau_0)}{T}), \tau_T).
\end{equation}
%and mitigating the peakiness effect.

In each decoding iteration, multiple edit operations occur, and each operation has a different decoding space size.
It may be beneficial to optimize this by using varying temperatures for each operation in every iteration. 
This is a complicated research question and we leave this exploration to future work.
%=================================================================================================
\section{Experiments}
\subsection{Experimental Setup}
\paragraph{Data \& Evaluation}
We use WMT'14 English-German (EN-DE)~\citep{bojar-etal-2014-findings} and WAT'17 English-Japanese (EN-JA) Small-NMT datasets~\citep{nakazawa-etal-2017-overview} for experiments.
We use BPE token-based BLEU scores for evaluations.
Data preprocessing follows~\citet{gu2019levenshtein}.

\paragraph{Baseline}
We use Levenshtein Transformer as our baseline.
Following~\citet{gu2019levenshtein}, we trained a LevT with 300K steps and a max batch size of 65,536 tokens per step.
However, like~\citet{reid2023diffuser}, we cannot reproduce the results of~\citet{gu2019levenshtein}.
We use our results in this paper.

\paragraph{RL} % 早すぎるアニーリングの結果は良くない
According to~\citet{gu2019levenshtein}, most decodings are gotten in 1-4 iterations, and the average number of decoding iterations is 2.43.
To minimize the gap between the training and decoding states, we start with a null string and conduct 3 iterations (8 edits) for each sample during RL training.
We set the total training steps $T$ to 50,000, with a max batch size of 4,096 tokens per step.
To prevent the out-of-memory issue, we limit the decoding space of placeholder insertion from 256 to 64.
The sample size $k$ of the baseline is set to 5.
Our implementation is based on Fairseq\footnote{\url{https://github.com/facebookresearch/fairseq}}.
\paragraph{Computational Cost}
The pre-training phase of LevT on a GCP VM instance with A100x4 GPUs requires roughly 3 days, while the subsequent RL fine-tuning process takes approximately 1 day to complete.
%-------------------------------------------------------------------------------------------------

\subsection{Results}
%-------------------------------------------------------------------------------------------------
% RLは効いている、即時報酬よりも累積報酬
% 即時報酬はなぜダメなのか、advantageなどを使って考察
We show the BLEU scores of our approaches in Table~\ref{tab:result}.
The episodic reward model\footnote{The term ``episode/stepwise reward model'' specifically refers to the model trained using the ``episode/stepwise reward maximization'' approach.} showed notable improvement over the baseline.
The score is even close to the distillation model, which requires a heavy pre-training\footnote{To produce a distillation model, we need to train an autoregressive Transformer first, which needs additional 3 days of training on our machine.} of AR models.
However, the stepwise reward model showed only limited improvement.
% To explain this, we show the standard deviation (SD) of the advantage of the stepwise reward model in Table~\ref{tab:std}, where the advantage is $r(y) - b(s)$ in Equation~\ref{rl}.
% The SD in the former part of edit operations is much larger than the latter part, which means the stepwise reward model focuses on learning the former part of edit operations and neglecting the latter part, resulting in biased learning and low performance.
% On the contrary, the episodic reward model uses the same reward for all edit operations, resulting in equal learning and better performance.
To explain this, we focus on the advantage, $r(y) - b(s)$, included in the policy gradient (Equation~\ref{rl}), as a larger value of the advantage can increase the policy gradient's magnitude.
A higher standard deviation (SD) of the advantages indicates larger fluctuations in policy gradients.
Table~\ref{tab:std} shows the SDs of the advantages of the stepwise reward model, with notably higher values in the early stages of edit operations compared to later stages.
This suggests that the stepwise reward model disproportionately focuses on early operations, potentially leading to uneven learning and reduced performance.
In contrast, the episodic reward model applies the same rewards and advantages across all operations, facilitating more uniform learning and improved performance.

We only report scores of applying RL to the model without distillation since we found that RL significantly improved the model without distillation (max 1.69 points) compared to when distillation was applied (max 0.5 point).
Moreover, when confronted with distillation models, it raises questions such as which data we should use for RL training, the original or the distillation one.
We leave these research questions to future work.
%-------------------------------------------------------------------------------------------------
\begin{table}[t]
\centering
\resizebox{0.8\columnwidth}{!}{%
\begin{tabular}{lcc}
\toprule
Model & EN-DE & EN-JA\\
\midrule
LevT &  24.03 & 31.76\\
LevT + distillation &  26.49 & - \\
\midrule
LevT + RL (stepwise) & 24.29 & 31.73\\
LevT + RL (episodic) & \textbf{25.72} & \textbf{32.75}\\
\bottomrule
\end{tabular}}
\caption{The BLEU scores of our approaches and the baseline. Temperatures are set to 1. Due to the limited computational resources, we only trained the distillation model for the EN-DE dataset using the ready-made distillation dataset.}
\label{tab:result}
\end{table}
%-------------------------------------------------------------------------------------------------
\begin{table}[t]
\centering
\resizebox{0.9\columnwidth}{!}{%
\begin{tabular}{cl|rr}
\toprule
Iteration & Edit Operation & EN-DE & EN-JA \\
\midrule
1 & Insert + Replace & 9.99 & 8.59\\
\midrule
2 & Delete  & 2.05 & 1.35\\
  & Insert + Replace & 3.28 &2.48\\
\midrule
3 & Delete & 1.67 & 1.29\\
  & Insert + Replace & 3.04 &1.60\\
\bottomrule
\end{tabular}}
\caption{Stepwise reward model's standard deviation (SD) of the advantage in each edit operation. Insertion and replacement share the same reward.}
\label{tab:std}
\end{table}
%-------------------------------------------------------------------------------------------------
\begin{table}[t]
\centering
\resizebox{0.86\columnwidth}{!}{%
\begin{tabular}{lcc}
\toprule
Temperature & EN-DE & EN-JA\\
\midrule
Constant ($\tau = 1$) & 25.72 & 32.75\\
Constant ($\tau = 0.5$) & \textbf{25.98} & 33.45\\
Constant ($\tau = 0.1$) & 25.76 & 33.60\\
\midrule
Annealing ($\tau = 1 \rightarrow 0.1$) & 25.83 & \textbf{33.76} \\
Annealing ($\tau = 0.1 \rightarrow 1$) & 25.90 & 33.43\\
% Annealing ($\tau = 0.1 \rightarrow 1 \rightarrow 0.1$) & \textbf{25.99} & \\
\bottomrule
\end{tabular}}
\caption{The BLEU scores of episodic reward models using different temperature settings.}
\label{tab:temp}
\end{table}
%-------------------------------------------------------------------------------------------------

We show the BLEU scores of different temperature settings in Table~\ref{tab:temp}.
Model performance varies significantly with temperature settings (max 1.01 points in EN-JA).
Among constant setting models, the model with a temperature of 0.5 performed best in EN-DE, and the model with a temperature of 0.1 performed best in EN-JA, indicating that too large temperature harms RL training.
The two models using annealing schedules performed great in both tasks, showing the effectiveness of the annealing algorithms for improving learning efficiency.
However, the annealing models did not always outperform the constant models, which suggests the difficulty of seeking the optimal temperature setting for NAR models' RL training.
Also, we found the inverted annealing model ({\small $\tau {=} 0.1 {\rightarrow} 1$}) begins dropping performance after 10,000 steps training in EN-JA, indicating that the speed of annealing will significantly affect the model training quality.

We also quickly surveyed the relationship between performance and the number of decoding iterations in RL.
The model performance dropped when we reduced the number of iterations to 2 during training and remained flat when we increased it to 4, indicating that our setting is reasonable.
%=================================================================================================
\section{Conclusion and Future Work}
% 自分で適用できるスケジューリング（エントロピーによる仕組み）
% 他のデータ、タスク
% NARモデルのRLの工夫
% 即時報酬の活用、PPO、ベースラインの作り方
% 多様性の分析
This paper explored the application of reinforcement learning to edit-based non-autoregressive neural machine translation.
By incorporating RL into the training process, we achieved a significant performance improvement.
By empirically comparing stepwise and episodic reward maximization, we analyzed the advantages and disadvantages of these RL approaches.
We plan to have a deeper exploration of stepwise reward maximization and find a way to alleviate training inequality for multiple edit operations in the future.

Our investigation of temperature settings in NAR softmax sampling provided insights into striking a balance between exploration and exploitation during training.
Although our annealing methods perform well, they are not optimal and still depend on manually adjusting the parameters such as total training steps.
In the future, we plan to develop a self-adaption temperature control method using various indicators like entropy and advantage SD.

The experiments in this paper focused on the basics, and we plan to do more study for practical applications in future work.
As our methods are orthogonal to existing research on NAR architectures, our next step involves exploring the methods' applicability across a broader spectrum, including state-of-the-art models.
Additionally, we plan to investigate how to effectively apply RL to the distillation model, the impact of different baseline designs on performance, and the impact of RL on output diversity.
% Using human feedback or LLM output for NAR training can also be a research direction.
Applying RL to NAR is a massive and complex research question.
We look forward to more researchers joining this topic.

\clearpage
\bibliography{anthology,custom}

\clearpage
\onecolumn
\appendix
\section{Pseudo code of stepwise reward maximization}
\label{pseudo-stepwise}
\begin{figure*}[h]
    \centering
    \includegraphics[width=0.91\linewidth]{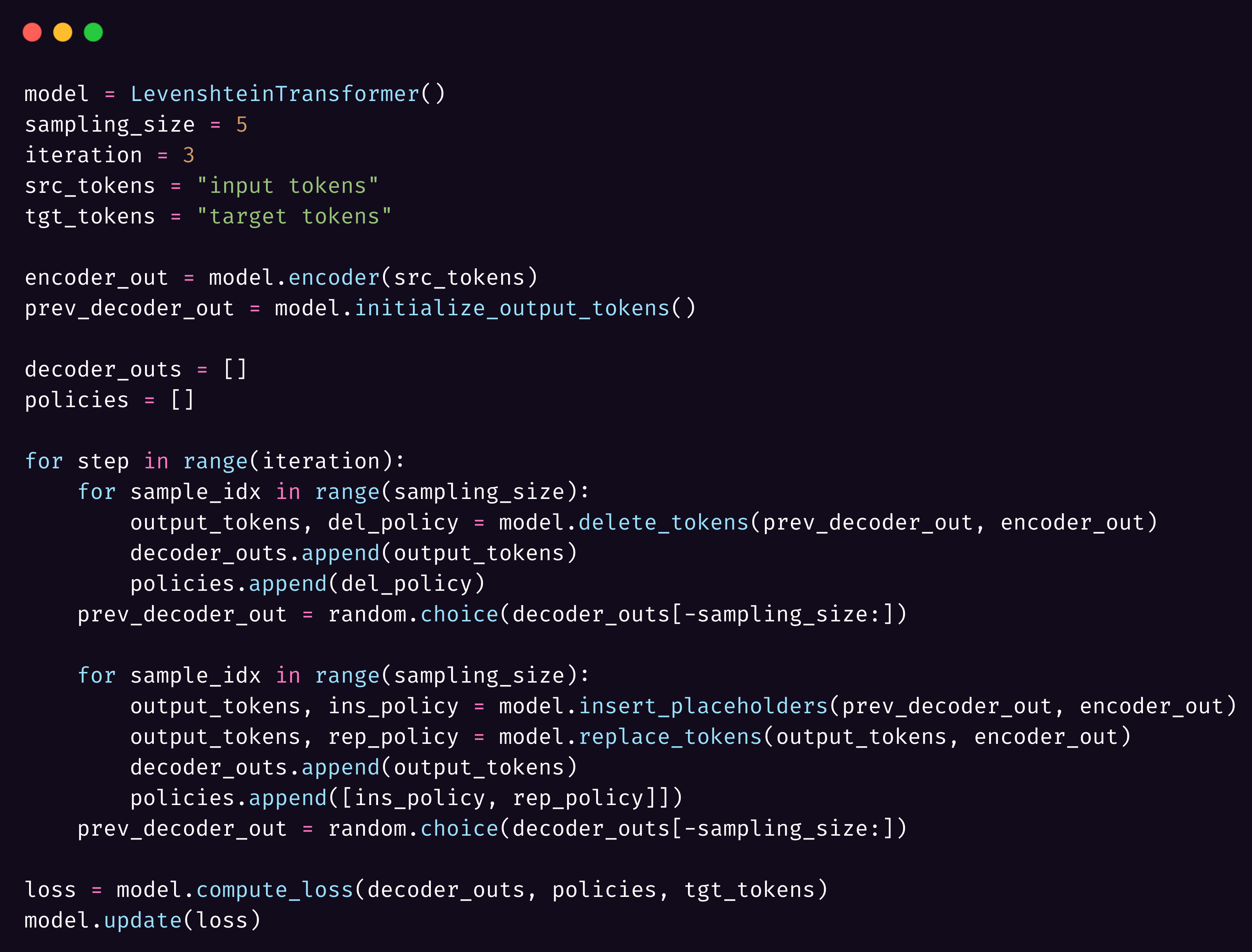}
    \caption{The pseudo code of stepwise reward maximization.}
    \label{fig:pseudo-fig-stepwise}
    \vspace{-10pt}
\end{figure*}

We show pseudo code of stepwise reward maximization in Figure~\ref{fig:pseudo-fig-stepwise}.

\section{Pseudo code of episodic reward maximization}
\label{pseudo-episodic}
\begin{figure*}[h]
    \centering
    \includegraphics[width=0.91\linewidth]{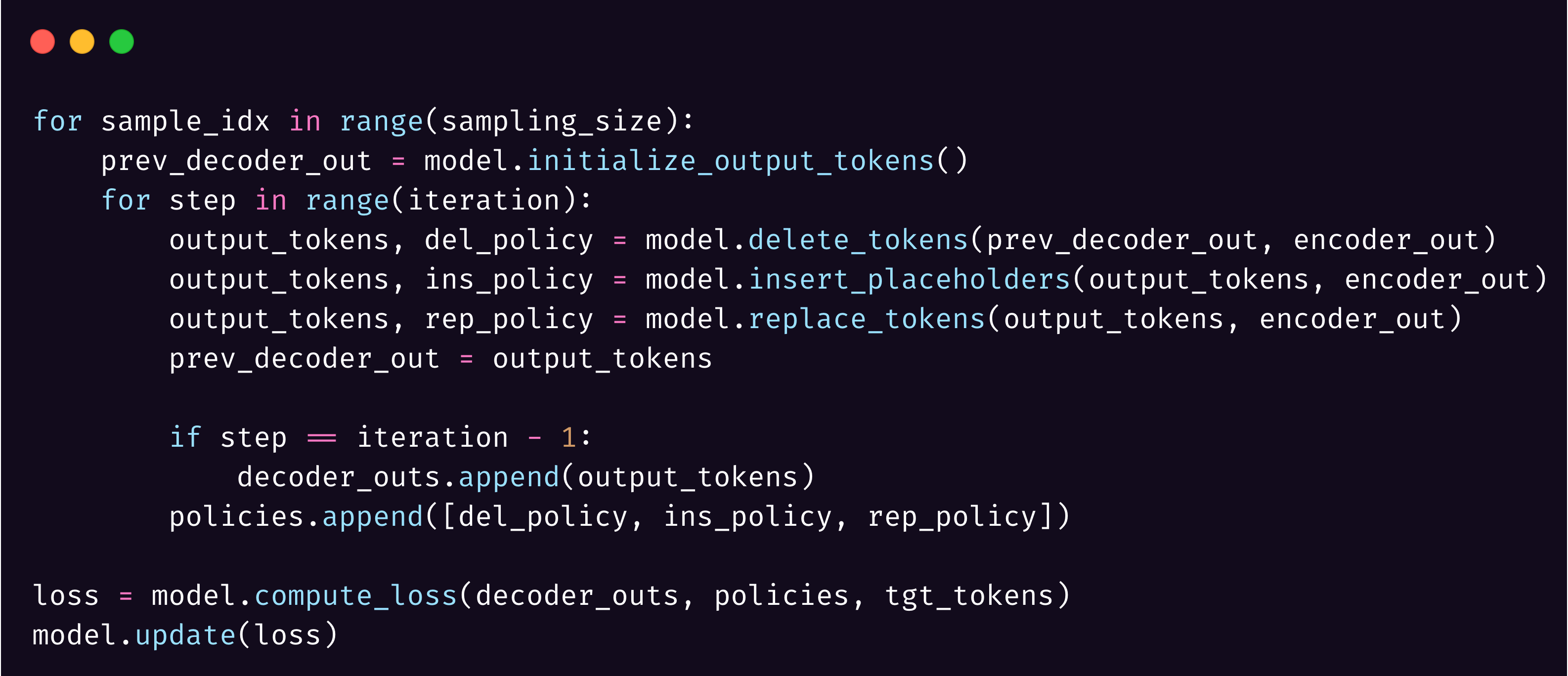}
    \caption{The pseudo code of episodic reward maximization.}
    \label{fig:pseudo-fig-episodic}
\end{figure*}

We show pseudo code of episodic reward maximization in Figure~\ref{fig:pseudo-fig-episodic}.

\end{document}